# Predictive Modeling for Breast Cancer Classification in the Context of Bangladeshi Patients: A Supervised Machine Learning Approach with Explainable AI

Taminul Islam [1], Md. Alif Sheakh [2], Mst. Sazia Tahosin [2], Most. Hasna Hena [2], Shopnil Akash [3], Yousef A. Bin Jardan [4], Gezahign FentahunWondmie [5,4*], Hiba-Allah Nafidi [6], Mohammed Bourhia [7*]

1. School of Computing, Southern Illinois University Carbondale, IL, United States
2. Department of Computer Science and Engineering, Daffodil International University, Dhaka, Bangladesh
3. Department of Pharmacy, Faculty of Allied Health Sciences, Daffodil International University, Dhaka, Bangladesh.
4. Department of Pharmaceutics, College of Pharmacy, King Saud University, P.O. Box 11451, Riyadh, Saudi Arabia.
5. Department of Biology, Bahir Dar University, P.O. Box 79, Bahir Dar, Ethiopia.
6. Department of Food Science, Faculty of Agricultural and Food Sciences, Laval University, 2325Quebec City, QC G1V 0A6, Canada.
7. Department of Chemistry and Biochemistry, Faculty of Medicine and Pharmacy, Ibn Zohr University, Laayoune 70000, Morocco.
Corresponding Author: resercherfent@gmail.com (GFW)

## Abstract

Breast cancer has rapidly increased in prevalence in recent years, making it one of the leading causes of mortality worldwide. Among all cancers, it is by far the most common. Diagnosing this illness manually requires significant time and expertise. Since detecting breast cancer is a time-consuming process, preventing its further spread can be aided by creating machine-based forecasts. Machine learning and Explainable AI are crucial in classification as they not only provide accurate predictions but also offer insights into how the model arrives at its decisions, aiding in the understanding and trustworthiness of the classification results. In this study, we evaluate and compare the classification accuracy, precision, recall, and F-1 scores of five different machine learning methods using a primary dataset (500 patients from Dhaka Medical College Hospital). Five different supervised machine learning techniques, including decision tree, random forest, logistic regression, naive bayes, and XGBoost, have been used to achieve optimal results on our dataset. Additionally, this study applied SHAP analysis to the XGBoost model to interpret the model's predictions and understand the impact of each feature on the model's output. We compared the accuracy with which several algorithms classified the data, as well as contrasted with other literature in this field. After final evaluation, this study found that XGBoost achieved the best model accuracy, which is 97%.

## 1. Introduction

Breast cancer begins when some cells in the breast start to grow uncontrollably, forming a mass called a tumor [1]. A breast cancer diagnosis typically falls into one of two main categories – benign (non-cancerous) or malignant (cancerous). Malignant tumors are dangerous as they can spread to distant sites in the body through the bloodstream or lymph system, a process known as metastasis [2,3]. Figure 1 (a) illustrates the distinction between benign and malignant tumors in terms of the normal cells and tumor cells, and (b) shows the benign and malignant masses. Benign tumors generally stay localized in one area and do not metastasize. Breast cancer manifests through several symptoms - a noticeable lump or mass in the breast, changes in breast size or shape compared to the other breast, alterations in the skin overlying the breast like dimpling or puckering, newly inverted nipple, redness or scaliness of breast skin, breast pain, and nipple discharge other than breast milk [4].

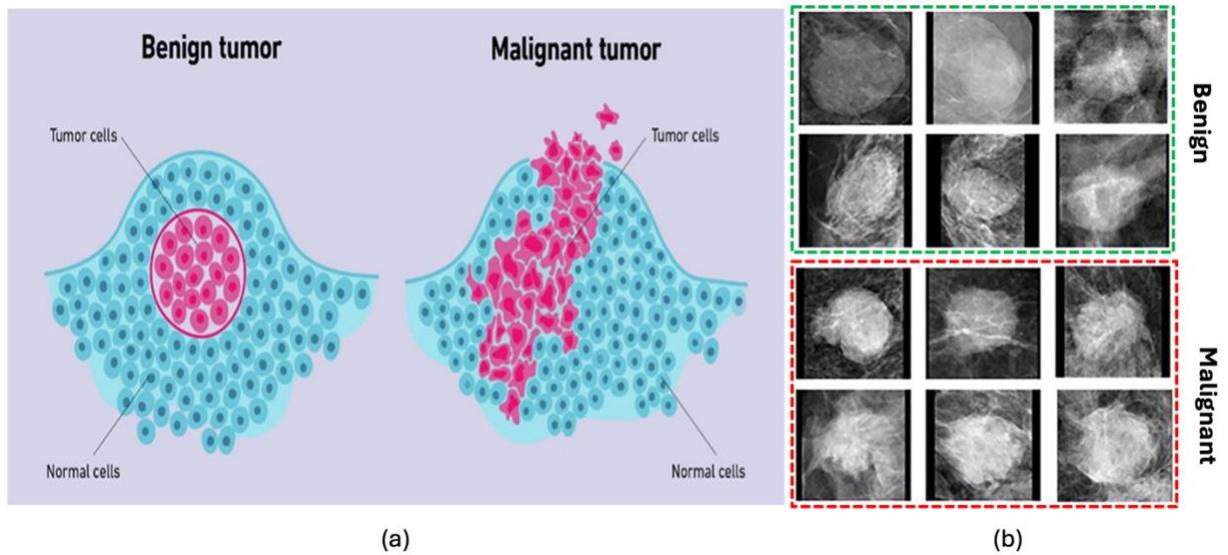

**Fig. 1.** Visualization of breast cancer: (a) benign and malignant tumor cells, (b) benign and malignant masses

Breast cancer is the second largest killer of women after cardiovascular disease. And more than 8% of women will experience it. Every year, more than 500,000 women are diagnosed with breast cancer, as stated in the World Health Organization's annual report [5]. In developing countries, due to a lack of screening programs and awareness, women often present at an advanced stage where treatment options are limited. Known risk factors for breast cancer include genetic mutations in BRCA genes, reproductive history (nulliparity, early menarche, late menopause), hormonal factors (use of hormone replacement therapy, oral contraceptives), obesity, alcohol consumption, smoking, radiation exposure at a young age and family history [6].

Many people were affected by cancer during this period. We can't pinpoint the origin of the sickness since it's tied to factors beyond our control. This is also a screening technique for identifying the cancer's aggressiveness. Several assessment items are connected to cancer detection, including clamp thickness, cell size consistency, and shape regularity. The outcome is difficult even for those tasked with inspiring others to take action, and yet the use of machine learning and other computer science techniques as general diagnostic tools has expanded in recent years. Countless numbers of people's lives have been saved by computer diagnostic programs that use diseases that have killed millions. In the realm of surgery, robotics is indispensable. Aside from other artificial intelligence's widespread usage in cancer detection, the system deployed in the intensive care unit is highly effective [7]. One in eight American females may get cancer between the ages of 15 and 19 [8]. Breast cancer is the result of unchecked cell division, which can also cause breasts to sag (called tumors) [9]. In most cases, the tumor poses no health risk. The necessity for precise categorization in the clinic may be a severe challenge for doctors and health workers, especially when the correct identification of the determinants might contribute to survival, regardless of whether the condition is benign or malignant.

The diagnosis of breast cancer involves a step-wise approach starting with a thorough clinical examination and radiological tests like mammograms and breast ultrasounds. This may be followed by tissue sampling through fine needle aspiration cytology (FNAC) or biopsy from suspicious areas and microscopic assessment to confirm malignancy [10]. As symptoms of breast cancer can be non-specific with wide variation across patients, the combination of these investigations is needed for accurate diagnosis in the majority of cases. So, to monitor and diagnose diseases, a human observer must be able to pick out very particular signal features. Due to the large number of patients in the critical care unit and the need for round-the-clock monitoring, several CAD approaches [11] for computer-aided medical systems have emerged in the recent decade to meet this issue. With these strategies, the challenge of classifying quantitative features may be posed rather than relying on qualitative diagnostic criteria. Machine learning algorithms can predict breast cancer diagnosis and prognosis [12]. The purpose of this work is to evaluate the efficacy and performance of these algorithms in terms of their accuracy, sensitivity, range, and precision. In the past 25 years, the importance of artificial intelligence has increased [13]. As scientists realize the importance of making firm decisions about how to treat certain diseases, the use of computers and machine learning as diagnostic tools has become deadly, which is the most serious disease screening task in the medical field [14]. One of the most important functions of disease is the definition of cancer. Using machine learning technology, doctors can detect, identify, and classify tumors as benign or malignant. There are some challenges in analyzing patient data and choosing doctors and specialists, but cognitive systems and computational methods (such as ML for classification) will ultimately help doctors and professionals [15]. However, as machine learning models become more complex, there is a need for Explainable AI (XAI) techniques to interpret these models and understand how they arrive at their predictions. Explainable AI methods like SHAP (SHapley Additive exPlanations) can shed light on how different features contribute to a model's output, increasing trust and transparency in the model's decision-making process [16]. In this study, we employ SHAP analysis on our best performing XGBoost model to explain its predictions and understand which factors have the greatest impact on determining if a patient has early-stage breast cancer or not. This study makes several key contributions to the prediction of early-stage breast cancer using supervised machine learning approaches:

- Employing hyperparameter tuning to optimize each machine learning algorithm and enhance performance.

- Utilizing a primary dataset for algorithm evaluation.

- Demonstrating that XGBoost achieved the highest accuracy of 97% and F1 score of 0.96, surpassing other algorithms.

- Conducting SHAP analysis on the XGBoost model to interpret its predictions and comprehend the impact of each feature on the model's output.

## 2. Literature review

Every day, the medical sector discovers new machine learning applications. The development is beneficial to scientific research. There is an abundance of research being conducted on this topic. Several research articles pertinent to our study have been uncovered. This project aims to provide a mechanism for predicting breast cancer. The majority of the dataset was obtained from the Dhaka Medical College Hospital. During this study, we were exposed to a few novel methodologies. Not a straightforward undertaking on our end. This notion will be discussed in further detail in the next chapter. To completely apply this study and to learn this new term, we examined prior research about the prediction of heart attacks.

Using their model, V. Chaurasia and T. Pal determined which machine learning algorithms performed the best in predicting breast cancer. In their study, they used Support Vector Machine (SVM), Naive Bayes (NB), Radial basis function Neural Networks (RBF NN), Decision Tree (DT), and a simplified version of Classification and Regression Trees (CART) [17]. After adopting their successful model, they obtained the highest Area Under the Curve (AUC) (96.84%) using Support Vector Machine on the original Wisconsin Breast Cancer datasets.

Djebbari et al. [18] evaluated if a machine learning ensemble might predict breast cancer survival time. Their breast cancer dataset they achieve a greater rate of accuracy using their technique than was seen in previous studies. S. Aruna and L. Nandakishore examine the performance of Decision Tree, Support Vector Machine, Naive Bayes, and K-Nearest Neighbors (K-NN) for classifying White Blood Cell (WBC)[19]Their top Support Vector Machine classifier AUC was 9 .6.99%.

M. Angrap used his six machine learning techniques to categorize tumor cells. Gated Recurrent Unit, a variant of the long short-term memory neural network, was created and implemented Gated recurrent unit (GRU). The SoftMax layer of the neural network was switched out for a layer of Support Vector Machine. With an accuracy of 99.04%, GRU Support Vector Machine performed best in that study [20]. Cross-validation was used by Karabatak et al. [21] to improve the accuracy of a model trained with association rules and a neural network to 95.6%. Naive Bayes classifiers were utilized, using a novel weight adjustment method.

Mohebian et al. [22] looked at the feasibility of using ensemble learning to foretell cancer recurrence. Three machine learning models that performed very well when fed a relevance vector were compared and contrasted by Gayathri et al. [23]. Payam et al. [24] employed a number of methods for preprocessing and data reduction, including a radial basis function network (RBFN), to achieve their goals.

Using information from breast cancer studies reported in [25], researchers created survival prediction models. In this study, we used survival prediction methods to both benign and malignant tumor of breast cancers. Extensive historical research shows that machine learning algorithms for breast cancer diagnosis have been investigated at length, as illustrated in [26]. They suggested that data augmentation strategies might help address the problem of having insufficient data. In [27], the authors showed how to automatically detect and identify cell structure using features of computer-aided mammography images. Many different methods of categorization and clustering have been evaluated, as reported in [28].

Fatih Muhammed and Ak [29] compared detection and diagnosis of breast cancer using data visualization and machine learning. Using Dr. William H. Walberg's breast tumor data, they used a variety of techniques including Logistic Regression (LR), nearest neighbor (NN), Support Vector Machine, simple Bayes, Decision Tree, random forest (RF), and convolutional forest using R, Minitab, and Python. Logistic regression with all features achieved the highest accuracy (98.1%), indicating high performance. Their research showed the benefits of data visualization and machine learning in cancer diagnosis, opening up new opportunities for cancer diagnosis.

Md. Islam et al. [30] compared five supervised machine learning methods for breast cancer prediction using the Wisconsin Breast Cancer Database. These methods include Support Vector Machine, Nearest Neighbor, Random Forest, Artificial neural networks (ANN), and Logistic regression. ANNs outperformed others by achieving the highest accuracy (98.57%), precision (97.82%) and F1 score (0.9890). The researchers concluded that machine learning for disease detection could provide medical staff with reliable and rapid responses to reduce the risk of death.

Vikas Chaurasia and Saurabh Pal [31] used machine learning to predict breast cancer using the Wisconsin Diagnostic Breast Cancer Database. They compared six algorithms, reduced features to 12 using statistical methods, and used ensemble methods to combine models. The results show that all the algorithms performed well, with a test accuracy of over 90%, especially in the refined feature section. His contributions include the use of feature selection and ensemble methods to improve breast cancer prediction accuracy.

Kabiraj et al. [32] Creating a breast cancer risk prediction model using Extreme Gradient Boosting (XGBoost) and Random Forest algorithms. The dataset used is from the UCI Machine Learning Repository. This approach includes the use of Random Forest and XGBoost methods, and the model achieves a classification accuracy of 74.73%.

Meerja Jabbar et al. [33] proposed a new ensemble method using Bayesian Network and Radial Basis function to classify breast cancer data. This method achieves 97% accuracy, better than existing approaches. The trial was conducted on the Wisconsin Breast Cancer Dataset (WBCD) using a variety of metrics to measure performance. The proposed ensemble study can help cancer specialists make accurate tumor diagnoses and support patients in making treatment decisions.

Shalini and Radhika [34] are working to predict breast cancer using different machine learning techniques. They use the UCI machine learning database and use artificial neural networks, Decision Tree, Support Vector Machine and Naive Bayes algorithms. As a result, a classification accuracy of 86% was found.

Naji et al. [35] Machine learning algorithms are used to predict and diagnose breast cancer. They compared five different algorithms, including Support Vector Machine, Random Forest, Logistic regression, Decision Tree (C4.5), and KNN, using the Wisconsin breast cancer diagnostic database. The main goal is to determine the best algorithm for breast cancer diagnosis. The results revealed that the support vector machine outperformed the other classifiers and achieved the highest accuracy of 97.2%. Research conducted with the Scikit learning library in the Anaconda Python environment contributes important insights to update breast cancer therapy and improve patient safety standards.

Puja Gupta and Shruti Garg [36] investigated breast cancer prognosis using six supervised machine learning algorithms and deep learning. The study includes a parametric analysis of each algorithm to achieve greater accuracy. The data set used in the study is not mentioned. The article describes data pre-processing, machine learning algorithms and their key parameters. The research results show that deep learning using Human Gradient Descent Learning is the most accurate with an accuracy rate of 98.24%. The paper concludes that proper hyperparameter machine learning tools can help identify tumors effectively.

**3. Methodology**

This research aims to anticipate breast cancer and achieve the highest level of precision possible. To run the model, we first build the dataset and choose which methods to employ. Supervised Learning [37] and Unsupervised Learning [38] are two techniques to develop a method in machine learning algorithms. We employed supervised machine learning methods in this study. In supervised learning, some classification methods are utilized to address classification issues. We employed the Decision Tree, Random Forest, Logistic Regression, Naïve Bayes, and XGBoost method in this study. In the forthcoming part on the suggested technique, we will explore all of the algorithms, how they function, and which one is the best, and we will attempt to determine which algorithm produces the best results based on the data set we gathered.

### 3.1. Dataset description

In this research, we have gathered a total of 500 patient's information from a government hospital in Bangladesh known as Dhaka Medical College Hospital. In this dataset, seven features were extracted from the image by Dhaka Medical College Hospital experts. The data has been used in this research with the consent of all patients. The dataset features are given below with a short description in Table 1. According to Fig 2, the data shows 254 noncancerous cases, while the remaining 246 cases are considered to be cancerous. Table 1, shows a short description of our dataset:

**Table 1.** Short description of our dataset.

| Label of the Dataset | Description |
| --- | --- |
| age | From 22 to 55 years women data is collected here. |
| mean_radius | The average of the radius of the tumor cells. |
| mean_texture | The average of the gray-scale values in the texture of the tumor cells. |
| mean_perimeter | The average size of the tumor cell's boundary. |
| mean_area | The average area of the tumor cells. |
| mean_smoothness | The average smoothness of the tumor cell's surface. |
| diagnosis | The classification of the breast tissue as benign is 0 or malignant is 1. |

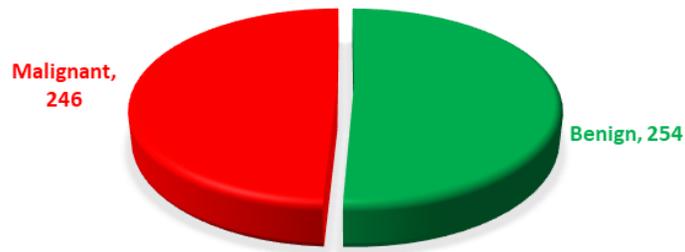

**Fig. 2.** Ratio of malignant and benign data based on overall dataset.

In this research we considered geometric features and age for the classification because geometric features, such as shape irregularity and size, play a crucial role in identifying abnormal growth patterns associated with malignant tumors. On the other hand, Age is a significant factor as breast cancer incidence increases with age, and the disease can manifest differently in younger versus older patients. By incorporating these features into our classification model, we aim to capture the distinct characteristics of benign and malignant tumors, improving the accuracy of our predictions. The full dataset was considered throughout the analysis of the dataset. It is seen in Figure 3 that the mean radius of the dataset is a counterpoint. Patients believed to have cancer have a radius bigger than 1, whereas those without symptoms have a radius nearer to 1. The full dataset was considered throughout the analysis of the dataset. It is seen in Figure 3 that the mean radius of the dataset is a counterpoint. Patients believed to have cancer have a radius bigger than 1, whereas those without symptoms have a radius nearer to 1.

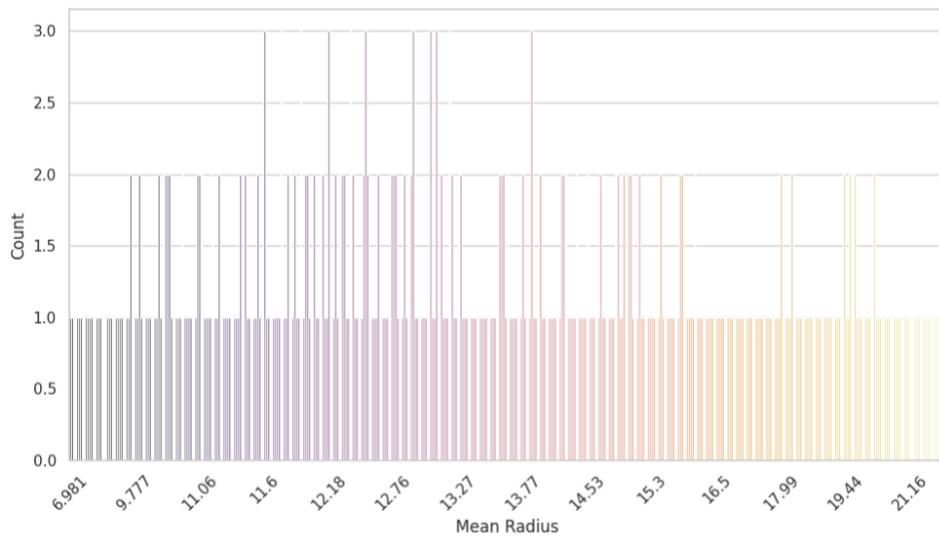

**Fig. 3.** Mean radius of this work on our dataset.

### 3.2. Proposed model workflow

The model workflow proposed in Figure 4 includes several steps, such as collecting the dataset, performing data preprocessing, splitting the data into 80% training and 20% testing sets, selecting the most relevant features, selecting a suitable supervised machine learning algorithm, classifying the samples into benign tumor or malignant tumor classes, and evaluating the model's performance. The initial step is to obtain the dataset needed to train the model. Once we obtain the dataset, we preprocess it by cleaning and translating the raw data into a machine-learning-friendly format. Following that, we divided the preprocessed dataset into two subsets: one for training and another for testing the model. After splitting, we select the most appropriate supervised machine learning method, which is a Random Forest classifier, Decision Tree, XGBoost, Naive Bayes, and Logistic Regression, and train this on the features. After training the model, we use it to categorize new samples into benign or malignant groups. Finally, we test the model's performance using several measures like AUC, precision, recall, $F_1$ score, and accuracy. During the training process, the model acquired the ability to identify and differentiate patterns and features that are indicative of malignant and benign instances of breast cancer, thereby enhancing its predictive capability. The performance of the model on the validation set was consistently monitored, and iterative fine-tuning was conducted until satisfactory results were achieved. After the model demonstrated robust predictive accuracy and generalization capability, we performed explainable AI using SHAP to interpret the model's predictions and understand the impact of each feature on the model's output. Additionally, we proceeded to implement the k-fold cross-validation technique. This proposed method ensures that the model is well-trained, correctly classifies samples, and performs optimally.

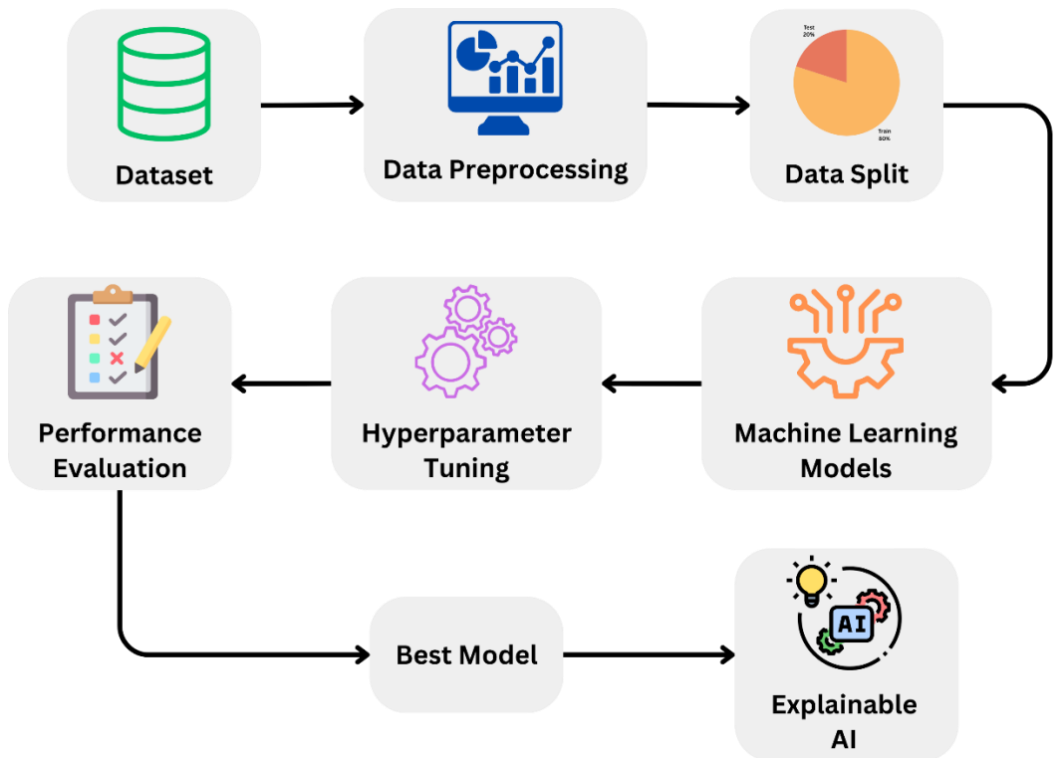

**Fig. 4.** Visualizing the workflow of the proposed model.

### 3.2.1. Data preprocessing

To make it suitable for machine learning we do cleaning and transforming raw data. This includes handling null values, scaling features using a standard scaler, encoding categorical variables, and normalizing data. The goal is to prepare data by improving the performance and efficiency of machine learning models.

### 3.2.2. Data splitting

After preprocessing we divided the dataset into a training set and a testing set. Here we use a common split ratio of 80:20, where 80% of the data is used to train the model and 20% is reserved for evaluating the model's performance. This ensures that the generalization ability of the model can be evaluated and prevents overfitting of the training data.

**Machine learning model**

Machine learning is the most practical way of predicting breast cancer sickness. Reading through the literature review, it becomes clear that the bulk of the work has been accomplished using machine learning and deep learning techniques. It is often understood that deep learning falls within the umbrella of machine learning. Five separate machine learning methods were used to this new dataset to find the most accurate method. Decision Tree, XGBoost, Logistic regression, Naive Bayes, and Random Forest are the categories used to organize these methods [39]. In this section, we'll get a brief overview of a few of these designs.

**Decision Tree (DT)**

Decision trees are widely used in machine learning for data classification and prediction. It can have different structures depending on the application. Although they work well for some classifiers, they can struggle with a large number of classes and limited training data. The resulting decision tree is easy to understand for domain experts, making it valuable for problem solving. In addition, it can be combined with ensemble methods to further improve performance. In summary, Decision trees are versatile and effective in many industries, including finance [40]. Decision trees classifier can be written as –

$$f_\Theta(x) = \sum_{\ell \epsilon leaves(\mathcal{T})} \theta_\ell \mathcal{T}_\ell(x) \qquad (1)$$

In equation 1, $f_\Theta(x)$ represents the model's final estimated for input vector x. $\mathcal{T}$ denotes the decision tree, $\theta_\ell$ is the weight associated with leaf node l, and $\mathcal{T}_\ell(x)$ denotes an indicator function that returns 1 if x falls within the region defined by leaf node l and 0 otherwise. The total of all the leaves in the tree ensures that the final prediction includes contributions from all of the different trees in the ensemble [41].

**Random Forest (RF)**

Random Forest is a popular machine-learning technique used for both classification and regression tasks. It creates multiple decision trees from different parts of the training data. Each tree classifies the data separately and the final prediction is the sum of all the individual forecasts. This approach reduces the risk of over fitting, leading to more accurate and reliable predictions. Additionally, as the number of trees increases, the method becomes more robust to noise and outliers in the data. However, there is a trade-off between accuracy and computational efficiency, as training more trees requires more time and resources. These algorithms divide the data recursively [42]. The Random Forest algorithm is provided below –

**Algorithm 1** Random Forest classifier

---

Node t, randomly selection v of the p independent variables.

∀of the k=1, …, vis the variable that was sampled; for each possible split of kth, determine the optimal split sk.

In s*, gain the optimal splitting k from k=1 to k=m; Determine the optimal split sk for splitting node t; jth variable is defined cut point cs* that is used for splitting node t.

The data is separated here, with the i=1, …, n; observation with xij<cs* going to the left descending node and all other observations going to the right descending node.

To grow a tree of maximum size Tb, simply repeat steps 1–4 for each node in the tree's descendent set.

---

**XGBoost (XGB)**

XGBoost (Extreme Gradient Boosting) is a widely used method for building robust predictive models, especially in machine learning [43]. It uses aggregated decision trees to make accurate predictions from complex datasets. Although decision trees are easy to interpret, XGBoost can be difficult to understand at first glance. However, data scientists and machine learning experts prefer XGBoost because it efficiently processes large datasets and quickly builds accurate models. This powerful and versatile tool strikes a balance between model complexity and prediction accuracy using gradient descent and regularization techniques. It is highly adaptable, making it valuable for extracting insights from complex datasets. However, careful optimization of the hyperparameters is necessary to achieve the best results. With the right approach, XGBoost enables data scientists to build accurate and reliable models that offer valuable insights into complex datasets [44]. XGBoost algorithm can be written as –

$$Obj = \sum_{i=1}^{m} l(y_i, \hat{y}_i) + \sum_{k=1}^{m} \Omega(f_k) \qquad (2)$$

Where $f_k$ is the leaf node's regular term of the kth classification tree, $l(y_i, \hat{y}_i)$ is the training error of sample xi, and Obj is the objective function [45].

### Naive Bayes (NB)

Naive Bayes is a classification algorithm that assumes that features are conditionally independent, given class labels. Although this assumption is often violated in real databases, it is still useful in practical applications. Although not independent, classification can derive information from features. For high-dimensional data, is fast and efficient with less training data than complicated and complex models because it estimates the probability of each feature separately. It can handle both discrete and continuous data, making it versatile for different databases. In natural language processing, it is essential for text categorization and spam filtering. Overall, Naive Bayes is a useful machine learning tool for solving classification problems [46]. Naive Bayes classifier can be written as –

$$f_c^{NB}(x) = \prod_{j=1}^{n} P(X_j = x_j | C = c) P(C = c) \tag{3}$$

In equation 3, Naive Bayes(x) is the probability that observation x is in class c, $f_c$ is the class-c observations, and n is the number f observations. $P(X_j = x_j | C = c)P(C = c)$ is the conditional probability of seeing feature j in class c. Divide class c's observations by feature j's $x_j$ usage. Training data can be used to calculate $P(C = c)$. Multiplying feature conditional probabilities and prior probabilities classifies new data into the most likely class [47].

### Logistic Regression (LR)

Using labeled data, we train our model in supervised learning. Logistic regression is used for categorization problems in supervised learning. Logistic regression's discrete output variable (y) is usually 0 or 1. A sigmoid function simulates X's effect on the output variable. This function provides a probability between 0 and 1 indicating the input's likelihood of being positive (1). Finance, marketing, and healthcare use logistic regression. Based on medical history and demographic data, logistic regression can estimate patients' cancer risk. Logistic regression handles nonlinear input-output relationships. For massive datasets, it requires fewer computing resources. Logistic regression's simplicity and effectiveness make it a common classification problem solution. Logistic regression can also provide relative relevance information for feature selection and model interpretation [48]. Logistic regression classifier can be written as –

$$g(x) = \ln\left(\frac{\pi(x)}{1-\pi(x)}\right) = \beta_0 + \beta_0 x_1 + \cdots + \beta_m x_m \quad (4)$$

Where $\pi(x)$ denotes the probability of a binary outcome (such as success or failure) given the values in the predictor vector x. The log odds are predicted as a linear combination of the predictor variables, and the coefficients $\beta_0, \beta_1, ..., \beta_m$ show the impacts of each predictor on the log chances. Exponentiating the equation allows one to determine the odds of a successful outcome given specific values of the predictors [49].

## 4. Experimental result

A lot of people make blunders in training or when extrapolating their results. Because the training error rate decreases with increasing model complexity, increasing the model's complexity can assist reduce training mistakes. The Bias-Variance Decomposition (Bias + Variance) method can be used to reduce the number of incorrect generalizations. Over fitting occurs when a reduction in training error results in an in-crease in test error rates. Each classification method may be judged by its accuracy, precision, recall, and F1 score.

When gauging the success of their models, writers used a wide range of techniques. While most studies looked at a combination of markers to determine how well they did, some just used one. The work is evaluated here using the criteria of accuracy, precision, recall, and the F1 score. For analyzing prediction data, this four-factor system is ideal. The capacity to appropriately recognize and categorize incidents is related to accuracy. Equation 5 [50] shows the formula of accuracy.

$$\text{Accuracy} = \frac{\text{True Positive} + \text{True Negative}}{\text{True Positive} + \text{False Positive} + \text{True Negative} + \text{False Negative}} \quad (5)$$

Specifically, accuracy in statistics is defined as the ratio of actual positive occurrences to the total predicted positive events. The mathematical expression of accuracy is given by Equation 6 [51].

$$\text{Precision} = \frac{\text{True Positive}}{\text{True Positive} + \text{False Positive}} \quad (6)$$

The term "harmonic mean" describes this method since it balances accuracy and memory. A version of the mathematical equation for the $F_1$ score is given by Equation 7 [52].

$$F_1 \ score = 2 \left(\frac{\text{Precision} \times \text{Recall}}{\text{Precision} + \text{Recall}}\right) \tag{7}$$

**4.1 Result analysis**

Table 2 provides information on hyperparameter tuning and each metric for different machine learning algorithms: Decision Tree, Random Forest, XGBoost, Naive Bayes, and Logistic regression. Hyperparameter tuning is an important step to improve the performance of machine learning models and involves finding the best combination of hyperparameters to achieve the highest precision, accuracy, recall, and $F_1$ score. Three hyperparameters are set for the decision tree algorithm: max_depth, min_samples_leaf, and min_samples_split. The best combination of hyperparameters that resulted in the highest $F_1$ score of 0.90 was a max_depth of 5, min_samples_leaf of 4, and min_samples_split of 5. Next, the Random Forest algorithm was tuned with four hyperparameters: max_depth, min_samples_leaf, min_samples_split, and n_estimators. The best combination of hyperparameters, which resulted in an impressive F1score of 0.94, included a min_samples_leaf of 1, min_samples_split of 5, and n_estimators of 300. The max_depth hyperparameter was not specified, indicating that the default value or automatic selection method might have been used. For the XGBoost algorithm, four hyperparameters were tuned: learning_rate, max_depth, n_estimators, and subsample. The best combination of hyperparameters achieved the highest $F_1$ score of 0.96, with a learning_rate of 0.01, max_depth of 3, n_estimators of 500, and subsample of 1.0. The Naive Bayes algorithm did not require hyperparameter tuning, and its default settings were used. It still achieved a respectable $F_1$ score of 0.94. Finally, the Logistic Regression algorithm was tuned for the hyperparameters regularization strength, max_iter, and penalty. The best combination of hyperparameters resulted in an $F_1$ score of 0.93, with a regularization strength of 10, max_iter of 100, and penalty using L2 regularization.

In terms of performance metrics, the XGBoost algorithm outperformed others with an effective precision of 0.97 and high precision, recall, and $F_1$ scores. The random forest algorithm also performed well with an accuracy of 0.96 and a balanced accuracy trade-off. The decision tree algorithm achieved good results with an accuracy of 0.91 and a balanced $F_1$ score of 0.90. Naive Bayes and Logistic regression show competitive performance with $F_1$ scores of 0.94 and 0.93, respectively. Overall, hyper-parameter tuning plays an important role in improving the performance of the model, and the choice of algorithm significantly influenced the final result, with XGBoost and Random Forest standing out as high-performance models.

**Table 2.** Hyperparameter tuning with performance metrics for all algorithms

| Algorithms | Hyperparameter Tuning | Range | Best | Accuracy | Precision | Recall | $F_1$ Score |
|---|---|---|---|---|---|---|---|
| Decision Tree | max_depth | None, 5, 10 | 5 | 0.91 | 0.94 | 0.89 | 0.9 |
| | min_samples_leaf | 2, 5, 10 | 4 | | | | |
| | min_samples_split | 1, 2, 2004 | 5 | | | | |
| Random Forest | max_depth | None, 5, 10, 20 | None | 0.96 | 0.93 | 0.95 | 0.94 |
| | min_samples_leaf | 1, 2, 2004 | 1 | | | | |
| | min_samples_split | 2, 5, 10 | 5 | | | | |
| | n_estimators | 100, 300, 500 | 300 | | | | |
| XGBoost | learning_rate | 0.01, 0.1, 0.3 | **0.01** | **0.97** | **0.94** | **0.95** | **0.96** |
| | max_depth | 3, 5, 2007 | **3** | | | | |
| | n_estimators | 100, 300, 500 | **500** | | | | |
| | subsample | 0.8, 1.0 | **1** | | | | |
| Naive Bayes | No | | | 0.94 | 0.99 | 0.9 | 0.94 |
| Logistic Regression | Regularization strength | 0.001, 0.01, 0.1, 1, 10 | 10 | 0.93 | 0.93 | 0.93 | 0.93 |

As can be seen in Table 2, it is evident that the accuracy of Random Forest and XGB is significantly greater than that of the other five machine-learning methods. When compared to the other algorithms that were used, the results of the Decision Tree method are significantly inferior to those of the Naive Bayes and Logistic Regression algorithms. In terms of the AUC comparison, the best results were achieved by Random Forest and XGB.

Confusion matrices (CM) can be used to rapidly and easily summarize a classification system's efficacy. When the quantity of observations across categories differs significantly, even when there are only two categories in the dataset, the categorization may be incorrect. For more insight into the precision of the classification approach, we can compute a CM (Figure 5, 6, 7, 8 and 9).

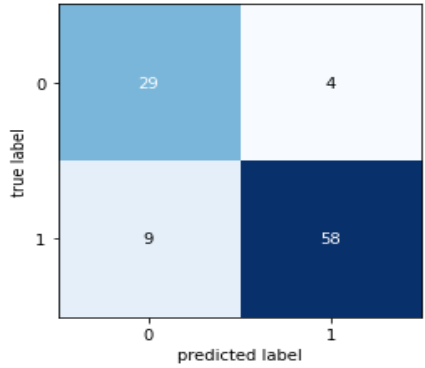

**Fig. 5.** CM of DT

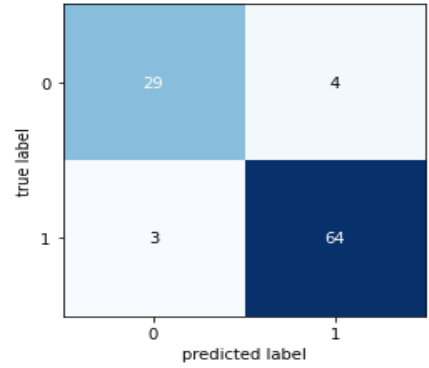

**Fig. 6.** CM of RF

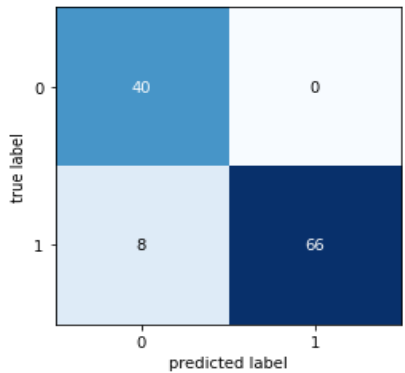

**Fig. 7.** CM of NB

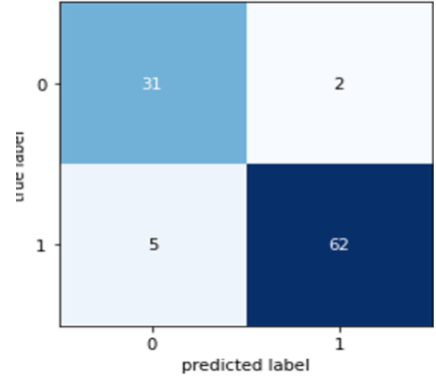

**Fig. 8.** CM of XGB

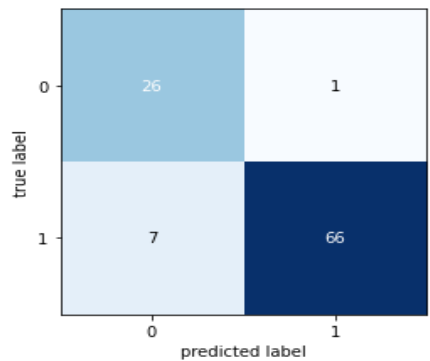

**Fig. 9.** CM of LR

Additionally, receiver operating characteristic (ROC) curves illustrate the diagnostic ability of a binary classifier as its discrimination threshold is varied. The area under the ROC curve (AUC) provides an aggregate measure across all possible classification thresholds. For our XGBoost model, the ROC AUC was 0.98, indicating excellent overall performance in distinguishing between the two classes shown in Figure 10. This high ROC AUC means the model is reliably assigning higher scores to positive instances than negative instances. We can have increased confidence in its ability to generalize well to new data. Further analysis into the confusion matrix for specific probability thresholds would provide deeper insight into preferred operating points along the ROC curve.

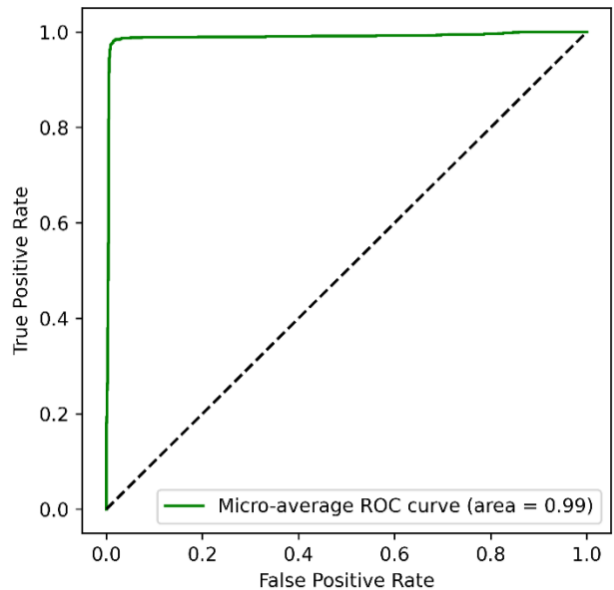

**Fig. 10.** ROC Curve of XGBoost Model

## 4.2 Performance Analysis Using Explainable AI

Here we perform SHAP analysis on our best performing XGBoost model.

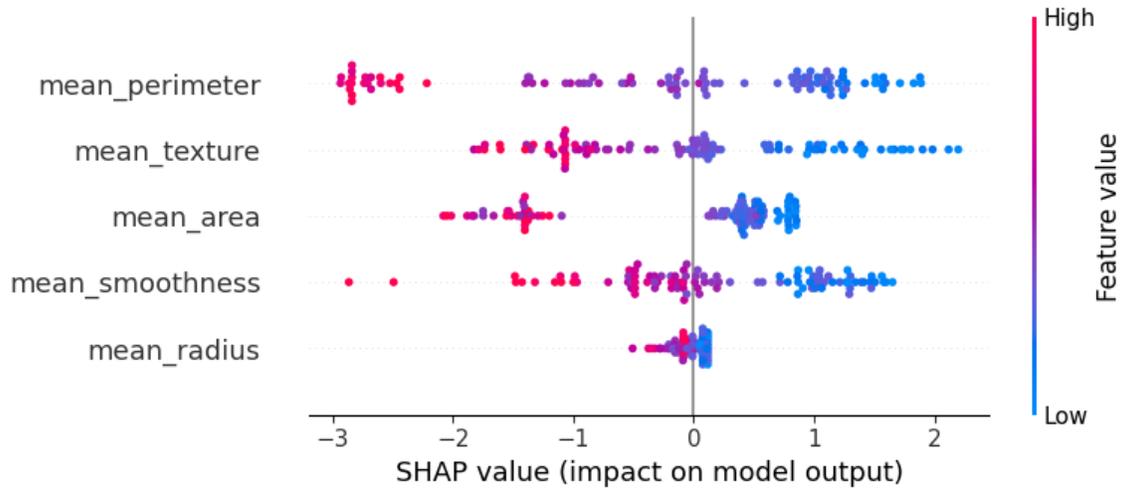

**Fig. 11.** SHAP Summary Plot for XGBoost Model

In Figure 11, the SHAP summary plot shows the impact of each feature on the model's output. The x-axis represents the SHAP value, where higher positive values indicate a higher probability of predicting early-stage breast cancer, and lower negative values indicate a lower probability. The features are ordered by their importance, with the most important features at the top. From the plot, we can observe that the mean_perimeter feature has the highest positive SHAP values, indicating that higher values of mean_perimeter contribute significantly to predicting early-stage breast cancer. On the other hand, the mean_radius feature has predominantly negative SHAP values, suggesting that lower values of mean_radius are associated with a higher likelihood of early-stage breast cancer.

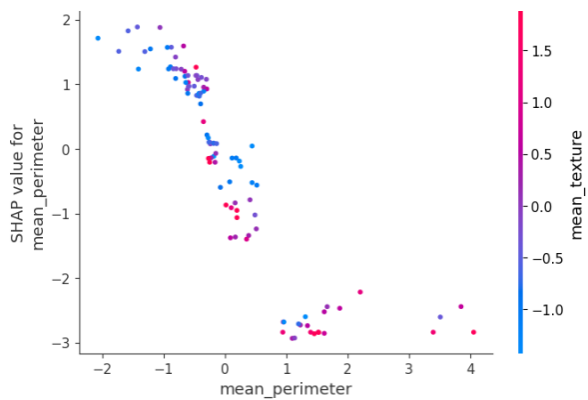

(a) mean_perimeter

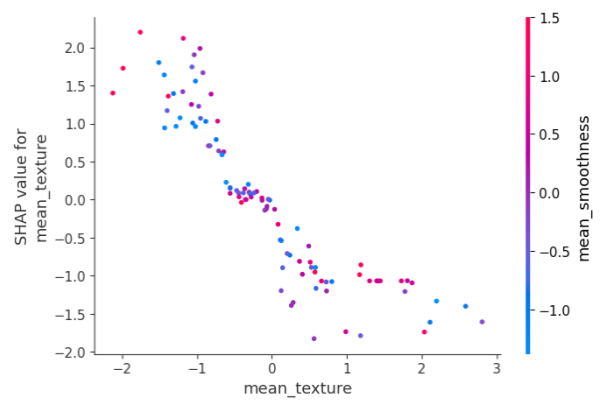

(b) mean_texture

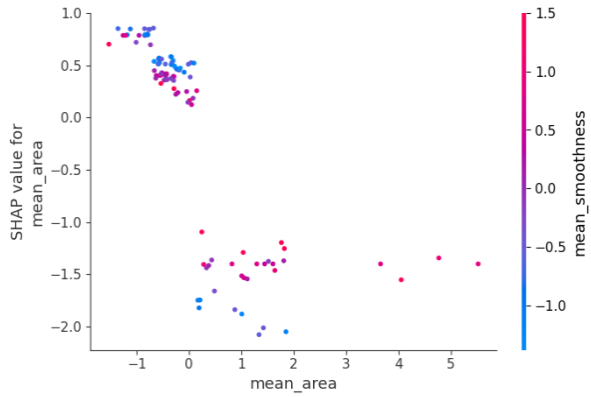

(c) mean_area

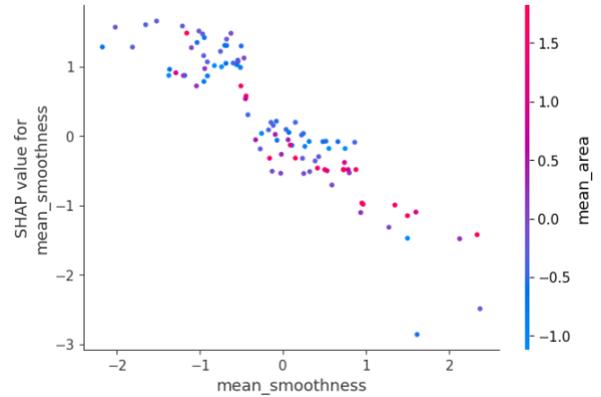

(d) mean_smoothness

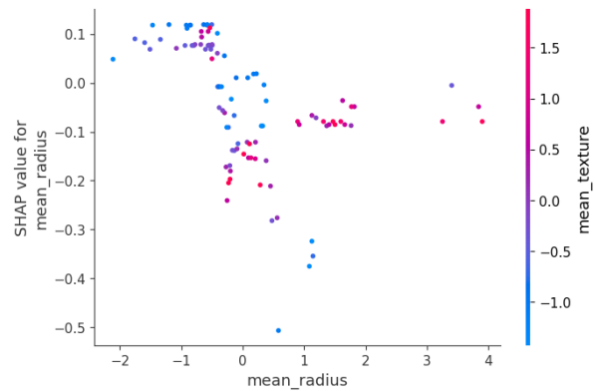

(e) mean_radius

**Fig. 12.** SHAP Dependence Plot for XGBoost Model

Figure 12 presents the SHAP dependence plots for various feature pairs, illustrating the relationship between the SHAP values and the feature values. In Figure 12(a), the dependence plot for mean_perimeter shows a weak positive correlation with the SHAP value, indicating that higher mean_perimeter values contribute to a higher probability of predicting early-stage breast cancer. The scatter plot for mean_smoothness and mean_texture in Figure 12(b) suggests a weak positive correlation, implying that as mean_smoothness increases, mean_texture also tends to increase. Figure 12(c) shows the plot for mean_area and mean_smoothness, which exhibits no clear linear relationship, with data points scattered throughout the area, suggesting no strong correlation or causation. Interestingly, the data points for mean_smoothness and mean_area in Figure 12(d) appear to exhibit a weak positive correlation, indicating that smoother surfaces tend to have larger areas, but with a fair amount of variation. Finally, Figure 12(e) displays the scatter plot for the

SHAP value and mean_radius, which shows no discernible relationship, with data points scattered throughout the area, suggesting no strong correlation or causation between these two variables.

**4.3 Performance Analysis Using Cross-validation**

When evaluating the transferability of statistical findings to a new dataset, researchers often employ a model validation technique known as cross-validation [53]. The result was calculated using K-fold cross-validation in this study. Using k-fold cross-validation, the dataset is split up into k smaller subsets. The remaining k1 subsets are combined for use as training samples, while the remaining subset is utilized to validate the others. The optimal value of k is dependent on the number of variables and the nature of the predictor, according to statistical theory. The sole adjustable aspect of the method is the number of subsamples (K) into which each data sample is divided. This method is typically referred to as k-fold cross-validation. For instance, k=10 would be referred to be 10-fold cross-validation if used in the model reference. Accuracy for all of the models in this study is shown in Table 3 using k-fold cross-validation.

**Table 3.** Evaluation of machine learning algorithms (accuracy) using k-fold cross-validation

| Algorithm | cv=10 | cv_score | cv_score (mean) |
|---|---|---|---|
| Decision Tree | 1 | 0.917718 | 0.907497 |
| | 2 | 0.923146 | |
| | 3 | 0.894783 | |
| | 4 | 0.899032 | |
| | 5 | 0.924925 | |
| | 6 | 0.90834 | |
| | 7 | 0.895156 | |
| | 8 | 0.913099 | |
| | 9 | 0.895104 | |
| | 10 | 0.903675 | |
| Random Forest | 1 | 0.919113 | 0.955505 |
| | 2 | 0.972991 | |

|  | 3 | 0.951697 |  |
|---|---|---|---|
|  | 4 | 0.940283 |  |
|  | 5 | 0.957985 |  |
|  | 6 | 0.95892 |  |
|  | 7 | 0.97818 |  |
|  | 8 | 0.957673 |  |
|  | 9 | 0.932415 |  |
|  | 10 | 0.981293 |  |
| XGBoost | 1 | 0.972403 | 0.973807 |
|  | 2 | 0.976671 |  |
|  | 3 | 0.98727 |  |
|  | 4 | 0.969338 |  |
|  | 5 | 0.967068 |  |
|  | 6 | 0.968395 |  |
|  | 7 | 0.963595 |  |
|  | 8 | 0.981956 |  |
|  | 9 | 0.988937 |  |
|  | 10 | 0.962441 |  |
| Naive Bayes | 1 | 0.915354 | 0.925651 |
|  | 2 | 0.917053 |  |
|  | 3 | 0.950712 |  |
|  | 4 | 0.914418 |  |
|  | 5 | 0.923784 |  |
|  | 6 | 0.908685 |  |
|  | 7 | 0.913652 |  |

| | 8 | 0.921779 | |
| | 9 | 0.949879 | |
| | 10 | 0.941254 | |
| Logistic Regression | 1 | 0.917948 | 0.927323 |
| | 2 | 0.917869 | |
| | 3 | 0.93195 | |
| | 4 | 0.929658 | |
| | 5 | 0.939511 | |
| | 6 | 0.924365 | |
| | 7 | 0.918985 | |
| | 8 | 0.935196 | |
| | 9 | 0.936036 | |
| | 10 | 0.921721 | |

In this study, five machine learning algorithms were implemented to determine the optimal model performance. Based on the findings presented in the results section, the XGBoost method generated the highest model accuracy of 97%. To assess the performance, the k-fold cross-validation technique was employed in this study. The detailed outcome can be observed in Table 3. The accuracy cross-validation revealed that five algorithms, namely Decision Tree, Random Forest, Logistic regression, Naive Bayes, and XGBoost, exhibited strong performance. A 90% mean accuracy score was obtained through 10-fold cross-validation for the Decision Tree model, while the Random Forest model achieved a 95% mean accuracy score. In contrast, XGBoost exhibits superior performance compared to alternative algorithms. A 10-fold cross-validation mean score of 97% was obtained from the algorithm, while Naive Bayes achieved a score of 92%. In the 10-fold cross-validation, Logistic regression achieved a mean score of 92%. The results of the accuracy 10-fold cross-validation score indicate that the quality of the model employed in this research is satisfactory.

## 5. Discussion

Accurately predicting breast cancer development is a critical objective in current research efforts. However, the current utilization of data in this particular field is still limited. ML techniques have been employed to improve the accuracy of cancer prediction. The efficiency of the categorization method employed in this study justifies its comparison with other research attempts to evaluate its public importance. The primary goal of this study was to determine the most effective machine learning methodologies for accurately evaluating breast cancer risk. It acknowledges the potentially surprising effects of the predictive capabilities within this domain.

While most research in this field focuses on a limited amount of publicly accessible datasets, resulting in comparable evaluations of algorithms, the current study aims to explore new data sources. Out of the five machine learning methods utilized in this dataset, XGBoost and RF demonstrated exceptional performance, achieving 97% and 96% accuracy. Significantly, XGBoost demonstrated superior effectiveness in handling a recently produced dataset. It is reasonable to consider that a more extensive and equally distributed dataset may result in even higher levels of precision in the long term. The Random Forest algorithm showed significant capability by achieving the second-highest accuracy in performance. The Random Forest algorithm demonstrates its superiority in managing intricate interactions and mitigating the issue of overfitting. On the other hand, XGBoost exhibits remarkable abilities in enhancing model performance by utilizing gradient-boosting approaches. Upon evaluating their respective performances within the context of our study, it is evident that both algorithms contribute substantially to the accuracy of predictions. The ability of Random Forest to effectively handle different data features is a valuable complement to XGBoost's capability to capture complicated patterns. These unique traits improve our comprehension of breast cancer prediction. However, it is essential to acknowledge that these methods have certain disadvantages, including sensitivity to noisy data and computational complexity. When considering the broader effects of a study, it is essential to acknowledge that the generalizability of the results may be influenced in ways such as the size and diversity of the dataset. Reliability could be enhanced by ensuring a more balanced and broader dataset. This discourse highlights the considerable prospects of these algorithms while also acknowledging their limitations and recognizing the need for further refinement of prediction models in Table 4.

**Table 4.** Comparative analysis of different published studies

| Ref | Main idea of the paper | Accuracy | Applied algorithms | Limitations |
|---|---|---|---|---|
| **This Work** | Proposed supervised machine learning algorithms to predict early-stage breast cancer | Decision Tree: 91%, Random Forest: 96%, XGBoost:97%, Naïve Bayes: 94%, Logistic Regression: 93% | Decision Tree Random Forest XGBoost Naïve Bayes Logistic Regression | Need more analysis with models. |
| 54 | The paper proposes an improved nine-layer convolutional neural network (CNN) for identifying abnormal breast in mammography using the open-access mini MIAS dataset. | The proposed method achieved a sensitivity of 93.4%, specificity of 94.6%, precision of 94.5%, and accuracy of 94.0% on the test set. | Convolutional neural network (CNN) | Could enhance the performance of the proposed method. |
| 55 | Proposed a parallel Bayesian hyperparameter to optimize stacked ensemble models for breast cancer survival prediction. | BSense model 83.9%, 87.3%, 91.1%, and 80.1% Area Under Curve (AUC) for TCGA, METABRIC, Metabolomics, and RNA-seq dataset, respectively. | Stacking of machine learning models, i.e., Deep Neural Network (DNN), Gradient Boosting Machine (GBM), and Distributed Random Forest (DRF) | Limited dataset |
| 56 | Outline a complete automated process using advanced computer techniques to identify and categorize structures within breast ultrasound images. | Achieved 91% accuracy in the classification. | Ensembles methods to combine the performance of the individual CNNs architectures. | Accuracy could be better |
| 57 | The paper focuses on using machine learning algorithms for breast cancer risk prediction and diagnosis. | SVM has the highest accuracy of 97.13% with the lowest error rate | Support Vector Machine (SVM), Decision Tree (C4.5), Naive Bayes (NB), and k Nearest Neighbors (k-NN) algorithms | Data Pre-Processing |
| 58 | The paper focuses on predicting breast cancer using machine learning models and varying parameters. | Deep learning using Adam Gradient Descent Learning is 98.24%. | k-Nearest Neighborhood, Logistic Regression, Decision Tree, Random Forest, Support Vector Machine | The provided sources do not mention any specific limitations of the paper. |

Table 4 summarizes several research papers related to the prediction and diagnosis of breast cancer using various machine learning and deep learning techniques. Each paper addresses different aspects of breast cancer detection and prediction, and they vary in terms of accuracy, applied algorithms, and limitations.

Yu et al. introduce a nine-layer convolutional neural network (CNN) for identifying abnormal breast tissue in mammography images. Their method achieves a sensitivity of 93.4% and specificity of 94.6%, showing good performance for image-based breast cancer detection. The limitation mentioned is a potential for further enhancing the method's performance. Parampreet et al. propose a Bayesian hyperparameter optimization technique for stacked ensemble models, achieving good Area under the Curve (AUC) scores for various datasets. They use machine learning models like Deep Neural Network (DNN), Gradient Boosting Machine (GBM), and Distributed Random Forest (DRF). The limitation here is the availability of a limited dataset. Alessandro et al. focus on automated image analysis of breast ultrasound images, achieving an accuracy of 91% using ensemble methods with CNN architectures. The limitation is the potential for improving accuracy. Hiba et al. primarily employ machine learning algorithms for breast cancer risk prediction and diagnosis, with Support Vector Machine (SVM) yielding the highest accuracy of 97.13%. They also use Decision Tree (C4.5), Naive Bayes (NB), and k-Nearest Neighbors (k-NN). A limitation mentioned is data pre-processing. Puja et al. use deep learning with Adam Gradient Descent Learning, achieving a high accuracy of 98.24%. They also employ various traditional machine learning models. The sources do not mention any specific limitations. Comparing this work to the others, the novelty lies in the application of multiple supervised machine learning algorithms for early-stage breast cancer prediction, providing a comprehensive approach. While the accuracy is competitive, the acknowledgment of potential for improvement shows a commitment to enhancing the model's performance. Additionally, by leveraging various algorithms, this research showcases versatility and adaptability, which can be advantageous in real-world clinical settings where different types of data may be available. This approach makes it a robust and promising breast cancer prediction and diagnosis solution. However, it's essential to continue refining the model to achieve even higher accuracy rates and overcome its limitations.

## 6. Limitations and Future Work

The study although presents a promising result, but it has several limitations. As we have already discussed, the dataset has used here was limited in size and diversity, potentially limiting the generalizability of the findings. Due to the sensitive nature of patient data and the challenges associated with data collection, we were constrained in the amount of data we could gather for this study. Additionally, the features provided in the dataset were extracted image properties rather than raw scans, restricting direct image-based analysis. In future we are aiming to collect a large number of patient dataset from Bangladesh and United States and applying our method to get the comparative result. We plan to collaborate with multiple healthcare institutions to gather a more extensive and diverse dataset, incorporating data from different demographics and regions to improve model generalizability. Additionally, we will explore newer machine learning algorithms, deep learning techniques, or ensemble methods to enhance the accuracy and efficiency of our models. Conducting in-depth analyses on feature importance and selection methods will provide valuable insights into the key factors influencing breast cancer prediction. Furthermore, we intend to validate our models in real clinical settings through prospective studies, working closely with healthcare professionals for validation.

## 6. Conclusion

This study demonstrates the potential of using supervised machine learning algorithms for early prediction of breast cancer. This research has collected 500 patients primary data from Dhaka Medical College Hospital and applied five supervised machine learning algorithms. In the evaluation this research found that XGBoost achieved the highest accuracy of 97%. XGBoost also achieved the highest precision (0.94), recall (0.95), and F1 score (0.96) rather than other algorithms. However, some limitations need to be acknowledged. More extensive real-world data is required to confirm the model's generalization capability across larger populations. Furthermore, the dataset only included derived picture attributes rather than raw scans, which limited the possibility of conducting direct image-based analysis. Despite these constraints, this work illustrates an important proof-of-concept for leveraging artificial intelligence to improve breast cancer diagnosis. The model can enable clinicians to rapidly screen patients and identify high-risk cases needing further examination. This would significantly impact early intervention and tailored treatment planning to improve survival outcomes. As next steps, integrating medical imagery into the pipeline and validating performance over larger multi-center datasets could help strengthen model robustness. It would also be valuable to experiment with combining machine predictions with expertise of oncologists to develop an augmented diagnostics system. Such human-AI collaboration can lead to more accurate and transparent cancer care.

**Author's contributions**

Conceptualization, writing the original draft, reviewing and editing: Taminul Islam, Md. Alif Sheakh, Mst. Sazia Tahosin. Formal analysis, investigations, funding acquisition, reviewing, and editing: Mst. Hasna Hena, Yousef A. Bin Jardan, Gezahign Fentahun Wondmie. Resources, data validation, data curation, and supervision: Hiba-Allah Nafidi, Mohammed Bourhia.

**Acknowledgments**

The authors would like to extend their sincere appreciation to the Researchers Supporting Project, King Saud University, Riyadh, Saudi Arabia for funding this work through the project number (RSP-2024R457)

**Ethical approval**

The current research is approved by the Dhaka Medical College Hospital Cancer Sample & Research Center Informed consent was obtained, and ethical guidelines were followed. The approval letter confirms that all necessary precautions were taken to ensure the protection of human subjects and adherence to ethical standards.

**Data Availability Statement**

The data that support the findings of this study are available from the corresponding author upon reasonable request.

**Conflict of Interest**: There is no Conflict of interest.